\pdfoutput=1
\documentclass[11pt]{article}
\usepackage{EMNLP2022}
\usepackage{hyperref}
\usepackage{pifont}
\newcommand{\cmark}{\ding{51}}
\usepackage{times}
\usepackage{latexsym}
\usepackage[T1]{fontenc}
\usepackage{multirow}
\usepackage[utf8]{inputenc}
\usepackage{microtype}
\usepackage{CJKutf8}
\usepackage{graphicx}
\usepackage{colortbl}
\definecolor{dGray}{gray}{.6}
\definecolor{mGray}{gray}{.9}

\title{It's about Time: Rethinking Evaluation on Rumor Detection Benchmarks using Chronological Splits}

\author{Yida Mu,  Kalina Bontcheva,  Nikolaos Aletras \\
  Department of Computer Science, The University of Sheffield \\
 \texttt{\{y.mu, k.bontcheva, n.aletras\}@sheffield.ac.uk} \\
  }

\begin{document}
\CJK{UTF8}{gbsn}
\maketitle
\begin{abstract}
New events emerge over time influencing the topics of rumors in social media. Current rumor detection benchmarks use random splits as training, development and test sets which typically results in topical overlaps. Consequently, models trained on random splits may not perform well on rumor classification on previously unseen topics due to the temporal concept drift. In this paper, we provide a re-evaluation of classification models on four popular rumor detection benchmarks considering chronological instead of random splits. Our experimental results show that the use of random splits can significantly overestimate predictive performance across all datasets and models. Therefore, we suggest that rumor detection models should always be evaluated using chronological splits for minimizing topical overlaps. 
\end{abstract}

\section{Introduction}

Unverified false rumors can spread faster than news from mainstream media, and often can disrupt the democratic process and increase hate speech \citep{vosoughi2018spread,zubiaga2018detection}. Automatic detection of rumors is an important task in computational social science, as it helps prevent the spread of false rumors at an early stage \citep{ma2017detect,zhou2019early,karmakharm-etal-2019-journalist,bian2020rumor}.  

Current rumor detection approaches typically rely on existing annotated benchmarks consisting of social media data, e.g., Twitter 15 \citep{ma2017detect}, Twitter 16 \citep{ma2017detect}, Weibo \citep{ma2016detecting}, and PHEME \citep{zubiaga2016analysing} that cover a wide range of time periods. These benchmarks use random splits for train, development and test sets which entail some topical overlap among them (see Table~\ref{tab:prev_work} for recent previous work). However, the distribution of topics in various NLP benchmarks (e.g., news, reviews, and biomedical) can be significantly affected by time \citep{huang2018examining,huang2019neural}. This is the phenomenon of temporal concept drift which can be induced by the changes in real-world events. Specifically, this also affects benchmarks on social media with new events such as elections, emergencies, pandemics, constantly creating new topics for discussion. 

\begin{table}[!t]
\resizebox{.47\textwidth}{!}{%
\centering
\begin{tabular}{|l|l|l|l|l|}

\hline
\rowcolor[HTML]{DAE8FC} \textbf{Paper}          & \textbf{Twitter 15}    & \textbf{Twitter 16}    & \textbf{PHEME}         & \textbf{Weibo}         \\ \hline

\multicolumn{1}{|c|}{\citet{tian2022duck}} & \multicolumn{1}{c|}{\cmark} & \multicolumn{1}{c|}{\cmark} & \multicolumn{1}{c|}{-} & \multicolumn{1}{c|}{\cmark} \\ \hline

\multicolumn{1}{|c|}{\citet{zeng2022early}} & \multicolumn{1}{c|}{-} & \multicolumn{1}{c|}{\cmark} & \multicolumn{1}{c|}{\cmark} & \multicolumn{1}{c|}{-} \\ \hline

\multicolumn{1}{|c|}{\citet{sheng2022zoom}} & \multicolumn{1}{c|}{-} & \multicolumn{1}{c|}{-} & \multicolumn{1}{c|}{-} & \multicolumn{1}{c|}{\cmark} \\ \hline

\multicolumn{1}{|c|}{\citet{mukherjee2022mtlts}} & \multicolumn{1}{c|}{-} & \multicolumn{1}{c|}{-} & \multicolumn{1}{c|}{\cmark} & \multicolumn{1}{c|}{-} \\ \hline

\multicolumn{1}{|c|}{\citet{sun2022rumor}} & \multicolumn{1}{c|}{\cmark} & \multicolumn{1}{c|}{\cmark} & \multicolumn{1}{c|}{\cmark} & \multicolumn{1}{c|}{-} \\ \hline

\multicolumn{1}{|c|}{\citet{de2021semantic}} & \multicolumn{1}{c|}{\cmark} & \multicolumn{1}{c|}{\cmark} & \multicolumn{1}{c|}{-} & \multicolumn{1}{c|}{-} \\ \hline

\multicolumn{1}{|c|}{\citet{ren2021cross}} & \multicolumn{1}{c|}{-} & \multicolumn{1}{c|}{-} & \multicolumn{1}{c|}{\cmark} & \multicolumn{1}{c|}{-} \\ \hline

\multicolumn{1}{|c|}{\citet{wei2021towards}} & \multicolumn{1}{c|}{\cmark} & \multicolumn{1}{c|}{\cmark} & \multicolumn{1}{c|}{\cmark} & \multicolumn{1}{c|}{-} \\ \hline

\multicolumn{1}{|c|}{\citet{li2021meet}} & \multicolumn{1}{c|}{-} & \multicolumn{1}{c|}{-} & \multicolumn{1}{c|}{\cmark} & \multicolumn{1}{c|}{-} \\ \hline

\multicolumn{1}{|c|}{\citet{rao-etal-2021-stanker}} & \multicolumn{1}{c|}{\cmark} & \multicolumn{1}{c|}{\cmark} & \multicolumn{1}{c|}{-} & \multicolumn{1}{c|}{\cmark} \\ \hline

\multicolumn{1}{|c|}{\citet{lin2021rumor}} & \multicolumn{1}{c|}{\cmark} & \multicolumn{1}{c|}{\cmark} & \multicolumn{1}{c|}{\cmark} & \multicolumn{1}{c|}{-} \\ \hline

\multicolumn{1}{|c|}{\citet{farinneya2021active}} & \multicolumn{1}{c|}{-} & \multicolumn{1}{c|}{-} & \multicolumn{1}{c|}{\cmark} & \multicolumn{1}{c|}{-} \\ \hline

\multicolumn{1}{|c|}{\citet{sun2021inconsistency}} & \multicolumn{1}{c|}{-} & \multicolumn{1}{c|}{-} & \multicolumn{1}{c|}{\cmark} & \multicolumn{1}{c|}{-} \\ \hline

\multicolumn{1}{|c|}{\citet{qian2021hierarchical}} & \multicolumn{1}{c|}{-} & \multicolumn{1}{c|}{-} & \multicolumn{1}{c|}{\cmark} & \multicolumn{1}{c|}{-} \\ \hline

\multicolumn{1}{|c|}{\citet{song2021adversary}} & \multicolumn{1}{c|}{\cmark} & \multicolumn{1}{c|}{\cmark} & \multicolumn{1}{c|}{\cmark} & \multicolumn{1}{c|}{-} \\ \hline

\multicolumn{1}{|c|}{\citet{kochkina2020estimating}} & \multicolumn{1}{c|}{\cmark} & \multicolumn{1}{c|}{\cmark} & \multicolumn{1}{c|}{\cmark} & \multicolumn{1}{c|}{-} \\ \hline

\multicolumn{1}{|c|}{\citet{yu2020coupled}} & \multicolumn{1}{c|}{-} & \multicolumn{1}{c|}{-} & \multicolumn{1}{c|}{\cmark} & \multicolumn{1}{c|}{-} \\ \hline

\multicolumn{1}{|c|}{\citet{xia2020state}} & \multicolumn{1}{c|}{-} & \multicolumn{1}{c|}{\cmark} & \multicolumn{1}{c|}{-} & \multicolumn{1}{c|}{\cmark} \\ \hline

\multicolumn{1}{|c|}{\citet{bian2020rumor}} & \multicolumn{1}{c|}{\cmark} & \multicolumn{1}{c|}{\cmark} & \multicolumn{1}{c|}{-} & \multicolumn{1}{c|}{\cmark} \\ \hline   

\multicolumn{1}{|c|}{\citet{lu2020gcan}} & \multicolumn{1}{c|}{\cmark} & \multicolumn{1}{c|}{\cmark} & \multicolumn{1}{c|}{-} & \multicolumn{1}{c|}{-} \\ \hline  
\end{tabular}
}
\caption{Recent work on rumor detection using random splits.}
\label{tab:prev_work}
\end{table}

\begin{table*}[!t]
\scriptsize
\centering
\begin{tabular}{|l|l|l|l|c|}
\hline
\rowcolor[HTML]{DAE8FC} \textbf{Dataset}                     & \textbf{id} & \textbf{Post}                       & \textbf{Label} & \multicolumn{1}{l|}{\textbf{Leven}} \\ \hline
\multirow{2}{*}{\textbf{Twitter 15}} &
   407231* &
  \begin{tabular}[c]{@{}l@{}}  r.i.p to the driver \textcolor{red}{who} died with paul walker that no one cares about because he wasn't famous.\end{tabular} &
  Rumor &
  \multirow{2}{*}{3} \\ \cline{2-4}
 &
  407236* &
  \begin{tabular}[c]{@{}l@{}}r.i.p to the driver \textcolor{blue}{that} died with paul walker that no one cares about because he wasn't famous.\end{tabular} &
  Rumor &
   \\ \hline
\multirow{2}{*}{\textbf{Twitter 16}} & 594687*     & the kissing islands, greenland. \textcolor{red}{URL} & Non-Rumor      & \multirow{2}{*}{0}                                 \\ \cline{2-4}
                                     & 604628*     & the kissing islands, greenland. \textcolor{blue}{URL} & Non-Rumor      &                                                    \\ \hline
\multirow{2}{*}{\textbf{PHEME}}      & 498483*     & \textcolor{red}{happening} now in \#ferguson URL     & Non-Rumor      & \multirow{2}{*}{9}                                 \\ \cline{2-4}
                                     & 499402*     & \textcolor{blue}{Right} now in \#ferguson URL         & Non-Rumor      &                                                    \\ \hline
\multirow{2}{*}{\textbf{Weibo}} &
  349863* &
  \begin{tabular}[c]{@{}l@{}}【喝易拉罐一定要吸管】一妇女喝了罐饮料，被送进医院，离开了世界。 
  研究显示罐上面的\\毒菌很多 \textcolor{red}{请转给你关心的朋友。} \textbf{Translation:} \textcolor{red}{Please forward to your friends you care about.} \end{tabular} &
  Rumor &
  \multirow{2}{*}{10} \\ \cline{2-4}
 &
  350023* &
  \begin{tabular}[c]{@{}l@{}}【喝易拉罐一定要吸管】一妇女喝了罐饮料，被送进医院，离开了世界。
研究显示罐上面的\\毒菌很多！！\textcolor{blue}{这些你知道么} \textbf{Translation:} \textcolor{blue}{Do you know about this?} \end{tabular} &
  Rumor &
   \\ \hline
\end{tabular}
\caption{Four pairs of posts from train and test data with similar or identical text content sampled from four rumor detection benchmarks. Post ids with close values indicate that two posts are published in the same period. \textbf{Leven} denotes the Levenshtein distance \citep{levenshtein1966binary} on character-level between the two posts with the same label (i.e., lower values indicate higher text similarity and vice versa). }
\label{tab:examples}
\end{table*}

\citet{gorman2019we} and \citet{sogaard2021we} have showed that using different data split strategies affects model performance in NLP downstream tasks.
Previous work has demonstrated that text classifiers performance significantly drops in settings where chronological data splits are used instead of random splits in various domains, e.g., hate speech, legal, politics, sentiment analysis, and biomedical \citep{huang2018examining,lukes2018sentiment,huang2019neural,florio2020time,chalkidis2022improved,agarwal2022temporal}. 
To minimize topical overlaps, a Leave-One-Out (LOO) evaluation protocol has been proposed~\citep{lukasik2015classifying,lukasik2016hawkes}. While this topic split strategy could potentially mitigate temporal concept drift, it still yields temporal overlaps between each subset and is practically not applicable to most common rumour detection benchmarks with a large number of topics (e.g., Twitter 15, Twitter 16, Weibo, etc.). We observe that the LOO protocol can be used for a few specific rumor detection benchmarks, such as (PHEME \citep{zubiaga2016analysing}), where each post is associated with a corresponding event, e.g, \textit{Ottawa Shooting} and \textit{Charlie Hebdo shooting}. 

Using random splits also results into posts with almost identical textual content shared during the same period. Table~\ref{tab:examples} displays four pairs of posts with \textbf{similar or identical} text content sampled from four different rumor detection benchmarks. This potential information leakage, results in classifying data almost identical to ones already being present in the training set. For practical application reasons, we believe that in order to evaluate a rumor detection system, it is necessary to detect not only long-standing rumors, but also emerging ones.




In this paper, we design a battery of controlled experiments to explore the hypothesis that whether temporality affects the predictive performance of rumor classifiers. To this end, we re-evaluate models on popular rumor detection benchmarks using chronological data splits i.e., by training the model with earlier posts and evaluating the model performance with the latest posts. 
Results show that the performance of rumor detection approaches trained with random data splits is significantly overestimated than chronological splits due to temporal concept drift. This suggests that rumor detection approaches should be evaluated with chronological data for real-world applications, i.e., to automatically detect emerging rumors.

\begin{table*}[!t]
\scriptsize
\centering
\begin{tabular}{|l|l|lll|ccc|lll|ccc|}
\hline
\rowcolor[HTML]{DAE8FC} 
\textbf{} &
  \multicolumn{1}{c|}{\cellcolor[HTML]{DAE8FC}\textbf{Benchmarks}} &
  \multicolumn{3}{c|}{\cellcolor[HTML]{DAE8FC}\textbf{Twitter 15}} &
  \multicolumn{3}{c|}{\cellcolor[HTML]{DAE8FC}\textbf{Twitter 16}} &
  \multicolumn{3}{c|}{\cellcolor[HTML]{DAE8FC}\textbf{PHEME}} &
  \multicolumn{3}{c|}{\cellcolor[HTML]{DAE8FC}\textbf{Weibo}} \\ \hline
\rowcolor[HTML]{DAE7EB} 
\multicolumn{1}{|c|}{\cellcolor[HTML]{DAE7EB}\textbf{Splits}} &
  \multicolumn{1}{c|}{\cellcolor[HTML]{DAE7EB}\textbf{Subsets}} &
  \multicolumn{1}{c|}{\cellcolor[HTML]{DAE7EB}Train} &
  \multicolumn{1}{c|}{\cellcolor[HTML]{DAE7EB}Dev} &
  \multicolumn{1}{c|}{\cellcolor[HTML]{DAE7EB}Test} &
  \multicolumn{1}{c|}{\cellcolor[HTML]{DAE7EB}Train} &
  \multicolumn{1}{c|}{\cellcolor[HTML]{DAE7EB}Dev} &
  Test &
  \multicolumn{1}{c|}{\cellcolor[HTML]{DAE7EB}Train} &
  \multicolumn{1}{c|}{\cellcolor[HTML]{DAE7EB}Dev} &
  \multicolumn{1}{c|}{\cellcolor[HTML]{DAE7EB}Test} &
  \multicolumn{1}{c|}{\cellcolor[HTML]{DAE7EB}Train} &
  \multicolumn{1}{c|}{\cellcolor[HTML]{DAE7EB}Dev} &
  Test \\ \hline
\rowcolor[HTML]{FFFFFF} 
\cellcolor[HTML]{FFFFFF} &
  \textbf{\# of Rumors} &
  \multicolumn{1}{l|}{\cellcolor[HTML]{FFFFFF}285} &
  \multicolumn{1}{l|}{\cellcolor[HTML]{FFFFFF}35} &
  52 &
  \multicolumn{1}{c|}{\cellcolor[HTML]{FFFFFF}-} &
  \multicolumn{1}{c|}{\cellcolor[HTML]{FFFFFF}-} &
  - &
  \multicolumn{1}{l|}{\cellcolor[HTML]{FFFFFF}1,420} &
  \multicolumn{1}{l|}{\cellcolor[HTML]{FFFFFF}72} &
  480 &
  \multicolumn{1}{c|}{\cellcolor[HTML]{FFFFFF}-} &
  \multicolumn{1}{c|}{\cellcolor[HTML]{FFFFFF}-} &
  - \\ \cline{2-14} 
\rowcolor[HTML]{FFFFFF} 
\multirow{-2}{*}{\cellcolor[HTML]{FFFFFF}\textbf{Standard Chronological}} &
  \textbf{\# of Non-Rumors} &
  \multicolumn{1}{l|}{\cellcolor[HTML]{FFFFFF}234} &
  \multicolumn{1}{l|}{\cellcolor[HTML]{FFFFFF}40} &
  96 &
  \multicolumn{1}{c|}{\cellcolor[HTML]{FFFFFF}-} &
  \multicolumn{1}{c|}{\cellcolor[HTML]{FFFFFF}-} &
  - &
  \multicolumn{1}{l|}{\cellcolor[HTML]{FFFFFF}2,641} &
  \multicolumn{1}{l|}{\cellcolor[HTML]{FFFFFF}508} &
  681 &
  \multicolumn{1}{c|}{\cellcolor[HTML]{FFFFFF}-} &
  \multicolumn{1}{c|}{\cellcolor[HTML]{FFFFFF}-} &
  - \\ \hline
\rowcolor[HTML]{FFFFFF} 
\cellcolor[HTML]{FFFFFF} &
  \textbf{\# of Rumors} &
  \multicolumn{1}{l|}{\cellcolor[HTML]{FFFFFF}260} &
  \multicolumn{1}{l|}{\cellcolor[HTML]{FFFFFF}37} &
  75 &
  \multicolumn{1}{l|}{\cellcolor[HTML]{FFFFFF}144} &
  \multicolumn{1}{l|}{\cellcolor[HTML]{FFFFFF}21} &
  \multicolumn{1}{l|}{\cellcolor[HTML]{FFFFFF}40} &
  \multicolumn{1}{l|}{\cellcolor[HTML]{FFFFFF}1,380} &
  \multicolumn{1}{l|}{\cellcolor[HTML]{FFFFFF}197} &
  394 &
  \multicolumn{1}{l|}{\cellcolor[HTML]{FFFFFF}1,645} &
  \multicolumn{1}{l|}{\cellcolor[HTML]{FFFFFF}235} &
  \multicolumn{1}{l|}{\cellcolor[HTML]{FFFFFF}470} \\ \cline{2-14} 
\rowcolor[HTML]{FFFFFF} 
\multirow{-2}{*}{\cellcolor[HTML]{FFFFFF}\textbf{\begin{tabular}[c]{@{}l@{}}Stratified Chronological\end{tabular}}} &
  \textbf{\# of Non-Rumors} &
  \multicolumn{1}{l|}{\cellcolor[HTML]{FFFFFF}259} &
  \multicolumn{1}{l|}{\cellcolor[HTML]{FFFFFF}37} &
  74 &
  \multicolumn{1}{l|}{\cellcolor[HTML]{FFFFFF}144} &
  \multicolumn{1}{l|}{\cellcolor[HTML]{FFFFFF}21} &
  \multicolumn{1}{l|}{\cellcolor[HTML]{FFFFFF}40} &
  \multicolumn{1}{l|}{\cellcolor[HTML]{FFFFFF}2,681} &
  \multicolumn{1}{l|}{\cellcolor[HTML]{FFFFFF}383} &
  766 &
  \multicolumn{1}{l|}{\cellcolor[HTML]{FFFFFF}1,619} &
  \multicolumn{1}{l|}{\cellcolor[HTML]{FFFFFF}231} &
  \multicolumn{1}{l|}{\cellcolor[HTML]{FFFFFF}463} \\ \hline
\rowcolor[HTML]{FFFFFF} 
\cellcolor[HTML]{FFFFFF} &
  \textbf{\# of Rumors} &
  \multicolumn{1}{l|}{\cellcolor[HTML]{FFFFFF}260} &
  \multicolumn{1}{l|}{\cellcolor[HTML]{FFFFFF}37} &
  75 &
  \multicolumn{1}{l|}{\cellcolor[HTML]{FFFFFF}144} &
  \multicolumn{1}{l|}{\cellcolor[HTML]{FFFFFF}21} &
  \multicolumn{1}{l|}{\cellcolor[HTML]{FFFFFF}40} &
  \multicolumn{1}{l|}{\cellcolor[HTML]{FFFFFF}1,380} &
  \multicolumn{1}{l|}{\cellcolor[HTML]{FFFFFF}197} &
  394 &
  \multicolumn{1}{l|}{\cellcolor[HTML]{FFFFFF}1,645} &
  \multicolumn{1}{l|}{\cellcolor[HTML]{FFFFFF}235} &
  \multicolumn{1}{l|}{\cellcolor[HTML]{FFFFFF}470} \\ \cline{2-14} 
\rowcolor[HTML]{FFFFFF} 
\multirow{-2}{*}{\cellcolor[HTML]{FFFFFF}\textbf{\begin{tabular}[c]{@{}l@{}}Random Splits\end{tabular}}} &
  \textbf{\# of Non-Rumors} &
  \multicolumn{1}{l|}{\cellcolor[HTML]{FFFFFF}259} &
  \multicolumn{1}{l|}{\cellcolor[HTML]{FFFFFF}37} &
  74 &
  \multicolumn{1}{l|}{\cellcolor[HTML]{FFFFFF}144} &
  \multicolumn{1}{l|}{\cellcolor[HTML]{FFFFFF}21} &
  \multicolumn{1}{l|}{\cellcolor[HTML]{FFFFFF}40} &
  \multicolumn{1}{l|}{\cellcolor[HTML]{FFFFFF}2,681} &
  \multicolumn{1}{l|}{\cellcolor[HTML]{FFFFFF}383} &
  766 &
  \multicolumn{1}{l|}{\cellcolor[HTML]{FFFFFF}1,619} &
  \multicolumn{1}{l|}{\cellcolor[HTML]{FFFFFF}231} &
  \multicolumn{1}{l|}{\cellcolor[HTML]{FFFFFF}463} \\ \hline  
\end{tabular}
\caption{Statistics of subsets. Note that using random splitting yields the same percentage of examples in each category as in the stratified chronological splits.}
\label{tab:statistics_subsets}
\end{table*}

\section{Methodology}

\subsection{Data}
We use four most popular rumor detection benchmarks, three in English and one in Chinese. Note that most related work is currently evaluating their rumor detection systems on two or three of these four benchmarks. (see Table~\ref{tab:prev_work}).
\paragraph{Twitter 15 and Twitter 16:} These datasets contain 1,490 and 818 tweets labeled into four categories including Non-rumor (NR), False Rumor (FR), True Rumor (TR), and Unverified Rumor (UR) introduced by \citet{ma2017detect}. 

    
\paragraph{PHEME:} This benchmark contains 5,802 verified tweets collected from 9 real-world breaking news events (e.g., Ottawa Shootting, Ferguson Unrest, etc.) associated with two labels, i.e., 1,972 Rumor and 3,830 Non-Rumor \citep{zubiaga2016analysing}. 
    
\paragraph{Weibo:} This dataset includes 4,664 verified posts in Chinese including 2,313 rumors debunked by the Weibo Rumor Debunk Platform\footnote{\url{https://service.account.weibo.com/?type=5&status=4}} and 2,351 non-Rumors from Chinese  media \citep{ma2016detecting}.

\paragraph{Data Pre-processing}
We opt for the binary setup (i.e., re-frame all benchmarks as rumor detection) to distinguish true/false information following \citet{lu2020gcan,rao-etal-2021-stanker}. We pre-process the posts by replacing @mention and hyperlinks with @USER and URL respectively. We also lowercase the tweets from three Twitter benchmarks.

\subsection{Data Splits}

\paragraph{Standard Chronological Splits}
For Twitter 15 and PHEME, we first sort all posts chronologically and then divide them into three subsets including a training set (70\% of the earliest data), a development set (10\% of data after train and before test), and a test set (20\% of the latest data). There is no temporal overlap between the three subsets.

\paragraph{Stratified Chronological Splits}
On the other hand, we observe that there is no temporal overlap between rumors and non-rumors in Twitter 16 and Weibo datasets. This suggests that it is not possible to use standard chronological splits as in Twitter 15 and PHEME. 

Therefore, we apply a \textbf{stratified chronological split} strategy for all benchmarks. We first split rumors and non-rumors separately in chronological order. We then divide them into three subsets (a total of six subsets), i.g., all rumors are split into a training set (70\% of the earliest rumors), a development set (10\% of data after train and before test), and a test set (20\% of the latest rumors). Finally, we merge the six subsets into the final three train, development and test sets. Note that this approach will result in no temporal overlap for \textbf{each label (i.e., rumor or non-rumor)} among the three final sets. We show the number of each split in Table \ref{tab:statistics_subsets}.



\paragraph{Random Splits}
Following standard practice (e.g., \citealt{bian2020rumor,lin2021rumor,rao-etal-2021-stanker}), we \textbf{randomly} split data using a 5-fold cross-validation. Note that these splits are made by preserving the percentage of posts in each category. Each split contains a training set (70\%), development set (10\%) and a test set (20\%) with the same ratio as in our chronological splits. 

\paragraph{Leave-One-Out (LOO) Splits}
For reference, we also provide the results of using the LOO evaluation protocol on PHEME dataset (see Table~\ref{tab:pheme_loo}).





\subsection{Models}
%
The main purpose of our experiments is to improve model evaluation by investigating the effects of temporal drifts in rumor detection by providing an extensive empirical study. Therefore, we opted using strong text classifiers that are generic and can be applied to all of our benchmarks:
\begin{itemize}
    \item \textbf{LR} We train a LR classifier using BOW to represent posts weighted by TF-IDF using a vocabulary of 5,000 n-grams. 
    \item \textbf{BERT} We directly fine-tune the BERT base model by adding a linear prediction layer on the top of the 12-layer transformer architecture following \citep{devlin2019bert}.    
    \item \textbf{BERT+ (BERTweet and ERNIE)} We also experiment with two domain specific models:  BERTweet \citep{bertweet} and ERNIE \citep{sun2020ernie} pre-trained on social media data using the same fine-tune strategy as the original BERT model.
\end{itemize}



\begin{table*}[!t]
\scriptsize
\centering
\begin{tabular}{|c|c|lll|lll|}
\hline
\rowcolor[HTML]{DAE8FC} 
\cellcolor[HTML]{DAE8FC} &
  \cellcolor[HTML]{DAE8FC} &
  \multicolumn{3}{c|}{\cellcolor[HTML]{DAE8FC}\textbf{Twitter15}} &
  \multicolumn{3}{c|}{\cellcolor[HTML]{DAE8FC}\textbf{PHEME}} \\ \cline{3-8} 
\rowcolor[HTML]{DAE8FC} 
\multirow{-2}{*}{\cellcolor[HTML]{DAE8FC}\textbf{Model}} &
  \multirow{-2}{*}{\cellcolor[HTML]{DAE8FC}\textbf{Strategy}} &
  \multicolumn{1}{c|}{\cellcolor[HTML]{DAE8FC}\textbf{P}} &
  \multicolumn{1}{c|}{\cellcolor[HTML]{DAE8FC}\textbf{R}} &
  \multicolumn{1}{c|}{\cellcolor[HTML]{DAE8FC}\textbf{F1}} &
  \multicolumn{1}{c|}{\cellcolor[HTML]{DAE8FC}\textbf{P}} &
  \multicolumn{1}{c|}{\cellcolor[HTML]{DAE8FC}\textbf{R}} &
  \multicolumn{1}{c|}{\cellcolor[HTML]{DAE8FC}\textbf{F1}} \\ \hline
 &
  \cellcolor[HTML]{C2F3D3}Random &
  \multicolumn{1}{l|}{\cellcolor[HTML]{C2F3D3}$86.7 \pm 2.1$} &
  \multicolumn{1}{l|}{\cellcolor[HTML]{C2F3D3}$85.2 \pm 1.8$} &
  \cellcolor[HTML]{C2F3D3}$85.0 \pm 1.8$ &
  \multicolumn{1}{l|}{\cellcolor[HTML]{C2F3D3}$84.1 \pm 1.2$} &
  \multicolumn{1}{l|}{\cellcolor[HTML]{C2F3D3}$79.3 \pm 1.0$} &
  \cellcolor[HTML]{C2F3D3}$80.9 \pm 1.0$ \\ \cline{2-8} 
 &
  Standard Chronological &
  \multicolumn{1}{l|}{$56.6 \pm 0.8$} &
  \multicolumn{1}{l|}{$56.3 \pm 0.7$} &
  $56.4 \pm 0.7$ &
  \multicolumn{1}{l|}{$67.3 \pm 0.1$} &
  \multicolumn{1}{l|}{$64.0 \pm 0.1$} &
  $63.9 \pm 0.1$ \\ \cline{2-8} 
\multirow{-3}{*}{\textbf{LR}} &
  Stratified Chronological &
  \multicolumn{1}{l|}{$56.3 \pm 2.5$} &
  \multicolumn{1}{l|}{$51.9 \pm 0.7$} &
  $41.4 \pm 0.4$ &
  \multicolumn{1}{l|}{$64.5 \pm 0.2$} &
  \multicolumn{1}{l|}{$63.0 \pm 0.3$} &
  $63.5 \pm 0.3$ \\ \hline
 &
  \cellcolor[HTML]{C2F3D3}Random &
  \multicolumn{1}{l|}{\cellcolor[HTML]{C2F3D3}$88.2 \pm 2.4$} &
  \multicolumn{1}{l|}{\cellcolor[HTML]{C2F3D3}$87.9 \pm 2.2$} &
  \cellcolor[HTML]{C2F3D3}$87.9 \pm 2.2$ &
  \multicolumn{1}{l|}{\cellcolor[HTML]{C2F3D3}$84.8 \pm 0.5$} &
  \multicolumn{1}{l|}{\cellcolor[HTML]{C2F3D3}$84.8 \pm 1.2$} &
  \cellcolor[HTML]{C2F3D3}$84.8 \pm 0.8$ \\ \cline{2-8} 
 &
  Standard Chronological &
  \multicolumn{1}{l|}{$54.8 \pm 4.0$} &
  \multicolumn{1}{l|}{$55.1 \pm 4.3$} &
  $52.9 \pm 3.6$ &
  \multicolumn{1}{l|}{$74.8 \pm 1.1$} &
  \multicolumn{1}{l|}{$75.1 \pm 0.8$} &
  $73.7 \pm 0.4$ \\ \cline{2-8} 
\multirow{-3}{*}{\textbf{BERT}} &
  Stratified Chronological &
  \multicolumn{1}{l|}{$58.2 \pm 7.3$} &
  \multicolumn{1}{l|}{$56.1 \pm 4.5$} &
  $52.8 \pm 5.6$ &
  \multicolumn{1}{l|}{$75.5 \pm 0.6$} &
  \multicolumn{1}{l|}{$77.7 \pm 0.5$} &
  $75.7 \pm 1.1$ \\ \hline
 &
  \cellcolor[HTML]{C2F3D3}Random &
  \multicolumn{1}{l|}{\cellcolor[HTML]{C2F3D3}$90.8 \pm 1.2$} &
  \multicolumn{1}{l|}{\cellcolor[HTML]{C2F3D3}$90.4 \pm 1.2$} &
  \cellcolor[HTML]{C2F3D3}$90.4 \pm 1.2$ &
  \multicolumn{1}{l|}{\cellcolor[HTML]{C2F3D3}$84.6 \pm 1.0$} &
  \multicolumn{1}{l|}{\cellcolor[HTML]{C2F3D3}$85.5 \pm 0.9$} &
  \cellcolor[HTML]{C2F3D3}$85.0 \pm 0.8$ \\ \cline{2-8} 
 &
  Standard Chronological &
  \multicolumn{1}{l|}{$58.6 \pm 1.9$} &
  \multicolumn{1}{l|}{$58.8 \pm 2.1$} &
  $57.4 \pm 2.5$ &
  \multicolumn{1}{l|}{$76.1\pm 1.1$} &
  \multicolumn{1}{l|}{$74.8 \pm 1.5$} &
  $71.6\pm 2.2$ \\ \cline{2-8} 
\multirow{-3}{*}{\textbf{BERT+}} &
  Stratified Chronological &
  \multicolumn{1}{l|}{$61.8 \pm 6.5$} &
  \multicolumn{1}{l|}{$57.9 \pm 2.4$} &
  $55.2 \pm 1.5$ &
  \multicolumn{1}{l|}{$75.3 \pm 0.9$} &
  \multicolumn{1}{l|}{$76.9 \pm 2.1$} &
  $71.0 \pm 3.5$ \\ \hline
\rowcolor[HTML]{DAE8FC} 
\cellcolor[HTML]{DAE8FC} &
  \cellcolor[HTML]{DAE8FC} &
  \multicolumn{3}{c|}{\cellcolor[HTML]{DAE8FC}\textbf{Twitter16}} &
  \multicolumn{3}{c|}{\cellcolor[HTML]{DAE8FC}\textbf{Weibo}} \\ \cline{3-8} 
\rowcolor[HTML]{DAE8FC} 
\multirow{-2}{*}{\cellcolor[HTML]{DAE8FC}\textbf{Model}} &
  \multirow{-2}{*}{\cellcolor[HTML]{DAE8FC}\textbf{Strategy}} &
  \multicolumn{1}{c|}{\cellcolor[HTML]{DAE8FC}\textbf{P}} &
  \multicolumn{1}{c|}{\cellcolor[HTML]{DAE8FC}\textbf{R}} &
  \multicolumn{1}{c|}{\cellcolor[HTML]{DAE8FC}\textbf{F1}} &
  \multicolumn{1}{c|}{\cellcolor[HTML]{DAE8FC}\textbf{P}} &
  \multicolumn{1}{c|}{\cellcolor[HTML]{DAE8FC}\textbf{R}} &
  \multicolumn{1}{c|}{\cellcolor[HTML]{DAE8FC}\textbf{F1}} \\ \hline
 &
  \cellcolor[HTML]{C2F3D3}Random &
  \multicolumn{1}{l|}{\cellcolor[HTML]{C2F3D3}$89.9 \pm 1.2$} &
  \multicolumn{1}{l|}{\cellcolor[HTML]{C2F3D3}$89.3 \pm 1.5$} &
  \cellcolor[HTML]{C2F3D3}$89.3 \pm 1.5$ &
  \multicolumn{1}{l|}{\cellcolor[HTML]{C2F3D3}$90.1 \pm 0.9$} &
  \multicolumn{1}{l|}{\cellcolor[HTML]{C2F3D3}$90.1 \pm 0.9$} &
  \cellcolor[HTML]{C2F3D3}$90.1 \pm 0.9$ \\ \cline{2-8} 
\multirow{-2}{*}{\textbf{LR}} &
  Stratified Chronological &
  \multicolumn{1}{l|}{$62.1 \pm 6.9$} &
  \multicolumn{1}{l|}{$55.8 \pm 4.7$} &
  $48.7 \pm 11.4$ &
  \multicolumn{1}{l|}{$79.1 \pm 0.1$} &
  \multicolumn{1}{l|}{$78.1 \pm 0.1$} &
  $77.9 \pm 0.1$ \\ \hline
 &
  \cellcolor[HTML]{C2F3D3}Random &
  \multicolumn{1}{l|}{\cellcolor[HTML]{C2F3D3}$91.9 \pm 1.0$} &
  \multicolumn{1}{l|}{\cellcolor[HTML]{C2F3D3}$91.5 \pm 0.8$} &
  \cellcolor[HTML]{C2F3D3}$91.5 \pm 0.8$ &
  \multicolumn{1}{l|}{\cellcolor[HTML]{C2F3D3}$92.3 \pm 1.2$} &
  \multicolumn{1}{l|}{\cellcolor[HTML]{C2F3D3}$92.2 \pm 1.2$} &
  \cellcolor[HTML]{C2F3D3}$91.2 \pm 1.2$ \\ \cline{2-8} 
\multirow{-2}{*}{\textbf{BERT}} &
  Stratified Chronological &
  \multicolumn{1}{l|}{$61.0 \pm 11.2$} &
  \multicolumn{1}{l|}{$54.3 \pm 4.3$} &
  $47.2 \pm 3.5$ &
  \multicolumn{1}{l|}{$89.0 \pm 2.5$} &
  \multicolumn{1}{l|}{$87.6 \pm 2.6$} &
  $87.5 \pm 2.6$ \\ \hline
 &
  \cellcolor[HTML]{C2F3D3}Random &
  \multicolumn{1}{l|}{\cellcolor[HTML]{C2F3D3}$89.8\pm 2.8$} &
  \multicolumn{1}{l|}{\cellcolor[HTML]{C2F3D3}$89.3 \pm 3.2$} &
  \cellcolor[HTML]{C2F3D3}$89.3 \pm 3.3$ &
  \multicolumn{1}{l|}{\cellcolor[HTML]{C2F3D3}$92.5 \pm .4$} &
  \multicolumn{1}{l|}{\cellcolor[HTML]{C2F3D3}$92.5 \pm .4$} &
  \cellcolor[HTML]{C2F3D3}$92.5 \pm .4$ \\ \cline{2-8} 
\multirow{-2}{*}{\textbf{BERT+}} &
  Stratified Chronological &
  \multicolumn{1}{l|}{$49.8 \pm 1.7$} &
  \multicolumn{1}{l|}{$49.9 \pm 0.9$} &
  $45.1 \pm 2.9$ &
  \multicolumn{1}{l|}{$88.1 \pm 2.5$} &
  \multicolumn{1}{l|}{$87.6 \pm 1.4$} &
  $88.5 \pm 1.5$ \\ \hline
\end{tabular}
\caption{Rumor detection prediction results across different data split methods. Green cells indicate that the model trained on random splits performs significantly better than both standard chronological splits and stratified chronological splits ($p<0.05$, t-test).} 
\label{tab:allresultssss}
\end{table*}

\begin{table}[!t]
\resizebox{\columnwidth}{!}{%
\begin{tabular}{|l|lll|}
\hline
\rowcolor[HTML]{DAE8FC} 
\multicolumn{1}{|c|}{\cellcolor[HTML]{DAE8FC}} & \multicolumn{3}{c|}{\cellcolor[HTML]{DAE8FC}\textbf{PHEME}}                   \\ \cline{2-4} 
\rowcolor[HTML]{DAE8FC} 
\multicolumn{1}{|c|}{\multirow{-2}{*}{\cellcolor[HTML]{DAE8FC}\textbf{Model}}} &
  \multicolumn{1}{c|}{\cellcolor[HTML]{DAE8FC}\textbf{P}} &
  \multicolumn{1}{c|}{\cellcolor[HTML]{DAE8FC}\textbf{R}} &
  \multicolumn{1}{c|}{\cellcolor[HTML]{DAE8FC}\textbf{F1}} \\ \hline
\textbf{LR}                                    & \multicolumn{1}{l|}{68.3 ± 3.8} & \multicolumn{1}{l|}{65.1 ± 6.3} & 63.2 ± 6.3 \\ \hline
\textbf{BERT}                                  & \multicolumn{1}{l|}{73.4 ± 3.1}  & \multicolumn{1}{l|}{71.9 ± 6.1}   & 70.7 ± 4.9  \\ \hline
\textbf{BERT+}                                 & \multicolumn{1}{l|}{75.3 ± 2.2}   & \multicolumn{1}{l|}{72.6 ± 8.1}   & 71.4 ± 7.0  \\ \hline
\end{tabular}%
}
\caption{Leave-One-Out evaluation protocol on PHEME dataset.}
\label{tab:pheme_loo}
\end{table}

\subsection{Hyperparameters and Implementation Details}
We train the model on the training set, perform model tuning and selecting on the development set, and evaluate performance on the test set. To evaluate the chronological data splits, we run the model five times with different random seeds for consistency. All chronological splits are available for reproducibility.\footnote{\url{https://github.com/YIDAMU/Rumor_Benchmarks_Temporality}}

For logistic regression, we use word-level and character-level tokenizers for Twitter and Weibo datasets respectively and only consider uni-gram, bi-grams, and tri-grams that appear in more than two posts for each dataset. For BERT, we set learning rate $lr = 2e-5$, batch size $bs = 32$, and maximum input length as 256 covering the max tokens of all posts. All BERT-style models are trained for 10 epochs using the early stopping method based on the loss on the development set. The best checkpoint model is saved for evaluation on the test set. The average run time of 10 epochs for the BERT model is less than 2 minutes. We employ Bert-Base-Uncased, Bertweet-Base and Chinese-Bert-WWM, Ernie-1.0 models from the HuggingFace library \citep{wolf2020transformers}. All experiments are conducted on a single NVIDIA V100 GPU with 32GB memory.

\subsection{Evaluation Metrics}
For all tasks, we report the averaged macro Precision, Recall and F1 values across five runs using different random seeds. 

\section{Results}
\paragraph{Random Splits vs. Chronological Splits}
Table~\ref{tab:allresultssss} shows the experimental results across all models and rumor detection benchmarks using \textbf{chronological splits} and random \textbf{5-fold cross-validation}. 
Overall, we observe that the use of random splits always leads to a significant overestimation of performance compared to chronological splits (t-test, $p<0.05$) across all models. Our results corroborate findings from previous work on studying temporal concept drift \citep{huang2018examining,chalkidis2022improved}. This suggests that chronological splits are necessary to more realistically evaluate rumor detection models.

We also note  that the effect of temporality varies in datasets of different size. For both data splitting strategies, we observe that the difference in performance is 50\% higher for the two datasets with hundreds of posts (e.g., Twitter 15 and Twitter 16) and around 10\% in ones with thousands of posts (e.g., PHEME and Weibo). 
For rumor detection tasks, temporality may have a greater impact on small-scale benchmarks than on large-scale benchmarks. 
For Twitter 16 and Weibo, the use of stratified chronological splits demonstrates significant performance drops compared to random splits due to the temporal concept drift. 


For chronological splits, we observe that pre-trained language models (i.e., BERT and BERT+) significantly outperform (t-test, $p<0.05$) logistic regression in all benchmarks. 
This is due to the fact that BERT-style models (i) outperform simpler linear models by a large margin in various NLP tasks~\cite{devlin2019bert}; and (ii) have been trained after the development of these four benchmarks implying some information leakage.


\begin{table*}[!t]
\centering
\scriptsize
\begin{tabular}{|cl|ccc|ccc|ccc|lcl|}
\hline
\multicolumn{2}{|c|}{\cellcolor[HTML]{DAE8FC}\textbf{Bechmark}} &
  \multicolumn{3}{c|}{\cellcolor[HTML]{DAE8FC}\textbf{Twitter 15}} &
  \multicolumn{3}{c|}{\cellcolor[HTML]{DAE8FC}\textbf{Twitter 16}} &
  \multicolumn{3}{c|}{\cellcolor[HTML]{DAE8FC}\textbf{PHEME}} &
  \multicolumn{3}{c|}{\cellcolor[HTML]{DAE8FC}\textbf{Weibo}} \\ \hline
\multicolumn{1}{|c|}{\textbf{Splits}} &
  \multicolumn{1}{c|}{\textbf{Test set}} &
  \multicolumn{1}{c|}{\textbf{total}} &
  \multicolumn{1}{c|}{\textbf{\#}} &
  \textbf{\%} &
  \multicolumn{1}{c|}{\textbf{total}} &
  \multicolumn{1}{c|}{\textbf{\#}} &
  \textbf{\%} &
  \multicolumn{1}{c|}{\textbf{total}} &
  \multicolumn{1}{c|}{\textbf{\#}} &
  \textbf{\%} &
  \multicolumn{1}{c|}{\textbf{total}} &
  \multicolumn{1}{c|}{\textbf{\#}} &
  \multicolumn{1}{c|}{\textbf{\%}} \\ \hline
\multicolumn{1}{|c|}{} &
  \ all posts &
  \multicolumn{1}{c|}{148} &
  \multicolumn{1}{c|}{3} &
  2\% &
  \multicolumn{1}{c|}{82} &
  \multicolumn{1}{c|}{6} &
  7\% &
  \multicolumn{1}{c|}{1161} &
  \multicolumn{1}{c|}{39} &
  3\% &
  \multicolumn{1}{l|}{933} &
  \multicolumn{1}{c|}{41} &
  4\% \\ \cline{2-14} 
\multicolumn{1}{|c|}{} &
  \# of wrong predictions&
  \multicolumn{1}{c|}{63} &
  \multicolumn{1}{c|}{2} &
  1\% &
  \multicolumn{1}{c|}{34} &
  \multicolumn{1}{c|}{2} &
  2\% &
  \multicolumn{1}{c|}{301} &
  \multicolumn{1}{c|}{5} &
  \textless{}1\% &
  \multicolumn{1}{l|}{99} &
  \multicolumn{1}{c|}{7} &
  \textless{}1\% \\ \cline{2-14} 
\multicolumn{1}{|c|}{\multirow{-3}{*}{\textbf{Chrono.}}} &
  \# of correct predictions&
  \multicolumn{1}{c|}{85} &
  \multicolumn{1}{c|}{1} &
  1\% &
  \multicolumn{1}{c|}{48} &
  \multicolumn{1}{c|}{4} &
  5\% &
  \multicolumn{1}{c|}{860} &
  \multicolumn{1}{c|}{34} &
  3\% &
  \multicolumn{1}{l|}{834} &
  \multicolumn{1}{c|}{34} &
  4\% \\ \hline
\multicolumn{1}{|c|}{} &
  all posts &
  \multicolumn{1}{c|}{\cellcolor[HTML]{C2F3D3}149} &
  \multicolumn{1}{c|}{\cellcolor[HTML]{C2F3D3}35} &
  \cellcolor[HTML]{C2F3D3}23\% &
  \multicolumn{1}{c|}{\cellcolor[HTML]{C2F3D3}83} &
  \multicolumn{1}{c|}{\cellcolor[HTML]{C2F3D3}26} &
  \cellcolor[HTML]{C2F3D3}30\% &
  \multicolumn{1}{c|}{\cellcolor[HTML]{C2F3D3}1161} &
  \multicolumn{1}{c|}{\cellcolor[HTML]{C2F3D3}181} &
  \cellcolor[HTML]{C2F3D3}16\% &
  \multicolumn{1}{l|}{\cellcolor[HTML]{C2F3D3}933} &
  \multicolumn{1}{c|}{\cellcolor[HTML]{C2F3D3}129} &
  \cellcolor[HTML]{C2F3D3}14\% \\ \cline{2-14} 
\multicolumn{1}{|c|}{} &
  \# of wrong predictions&
  \multicolumn{1}{c|}{\cellcolor[HTML]{C2F3D3}12} &
  \multicolumn{1}{c|}{\cellcolor[HTML]{C2F3D3}0} &
  \cellcolor[HTML]{C2F3D3}0\% &
  \multicolumn{1}{c|}{\cellcolor[HTML]{C2F3D3}5} &
  \multicolumn{1}{c|}{\cellcolor[HTML]{C2F3D3}1} &
  \cellcolor[HTML]{C2F3D3}1\% &
  \multicolumn{1}{c|}{\cellcolor[HTML]{C2F3D3}150} &
  \multicolumn{1}{c|}{\cellcolor[HTML]{C2F3D3}14} &
  \cellcolor[HTML]{C2F3D3}1\% &
  \multicolumn{1}{l|}{\cellcolor[HTML]{C2F3D3}65} &
  \multicolumn{1}{c|}{\cellcolor[HTML]{C2F3D3}4} &
  \cellcolor[HTML]{C2F3D3}\textless{}1\% \\ \cline{2-14} 
\multicolumn{1}{|c|}{\multirow{-3}{*}{\textbf{Random}}} &
  \# of correct predictions &
  \multicolumn{1}{c|}{\cellcolor[HTML]{C2F3D3}137} &
  \multicolumn{1}{c|}{\cellcolor[HTML]{C2F3D3}35} &
  \cellcolor[HTML]{C2F3D3}23\% &
  \multicolumn{1}{c|}{\cellcolor[HTML]{C2F3D3}78} &
  \multicolumn{1}{c|}{\cellcolor[HTML]{C2F3D3}25} &
  \cellcolor[HTML]{C2F3D3}30\% &
  \multicolumn{1}{c|}{\cellcolor[HTML]{C2F3D3}1011} &
  \multicolumn{1}{c|}{\cellcolor[HTML]{C2F3D3}167} &
  \cellcolor[HTML]{C2F3D3}14\% &
  \multicolumn{1}{l|}{\cellcolor[HTML]{C2F3D3}868} &
  \multicolumn{1}{c|}{\cellcolor[HTML]{C2F3D3}125} &
  \cellcolor[HTML]{C2F3D3}14\% \\ \hline
\end{tabular}
\caption{Error Analysis for all benchmarks. \# denotes the number of posts that are similar to posts from training set, i.e., known data. \% denote the percentage of similar posts in the test set. We set the threshold value to 20, which indicates that there are two or three different words between the two tweets.}
\label{tab:error}
\end{table*}

\begin{table*}[t]
\scriptsize
\centering
\begin{tabular}{|l|l|c|c|c|c|}
\hline
\rowcolor[HTML]{DAE8FC} 
 &
  \multicolumn{1}{c|}{\cellcolor[HTML]{DAE8FC}\textbf{Example}} &
  \multicolumn{1}{c|}{\cellcolor[HTML]{DAE8FC}\textbf{Test}} &
  \multicolumn{1}{c|}{\cellcolor[HTML]{DAE8FC}\textbf{Train}} &
  \multicolumn{1}{c|}{\cellcolor[HTML]{DAE8FC}\textbf{Correct}} &
  \multicolumn{1}{c|}{\cellcolor[HTML]{DAE8FC}\textbf{Wrong}} \\ \hline
\textbf{Twitter 15} &
  \begin{tabular}[c]{@{}l@{}}\#rip to the driver who died with \#paulwalker that no one cares about because he wasn't famous.\end{tabular} &
  4 &
  6 &
  4 &
  0 \\ \hline
\textbf{Twitter 16} &
  \begin{tabular}[c]{@{}l@{}}steve jobs was adopted. his biological father was abdulfattah jandali, a syrian muslim\end{tabular} &
  2 &
  13 &
  2 &
  0 \\ \hline
\textbf{PHEME} &
  Police are leaving now . \#ferguson HTTPURL &
  4 &
  11 &
  4 &
  0 \\ \hline
\multicolumn{1}{|c|}{\textbf{Weibo}} &
  \begin{tabular}[c]{@{}l@{}}【交通新规】2013年1月1日施行:1... 扩散给大家! 「广州日报」\\ \textbf{Translation:} {[}New driving laws{]} From 1 Jan 2013: Running a red light will result in a fine of\\ 100 RMB and 6 points. ... Spread the news to everyone! {[}Guangzhou Daily{]}\end{tabular} &
  2 &
  6 &
  2 &
  0 \\ \hline
\end{tabular}
\caption{Four examples of correct predictions using random splits, which artificially removes temporal concept drift. For example, in Twitter 15, there are 4 and 6 similar posts about rumors related to Paul Walker in the test set and the training set respectively.}
\label{tab:error_2}
\end{table*}

\paragraph{Standard vs. Stratifield Chronological Splits}
Note that dividing the datasets into standard chronological splits results in subsets that do not preserve the sample percentages for each category (see Table~\ref{tab:statistics_subsets}).
The upper part of Table~\ref{tab:allresultssss} displays the difference in model performance between two types of chronological splits on Twitter 15 and PHEME.
We observe that using both standard and stratified chronological splits results in similar model predictive performance (t-test, $p>0.05$). Even though stratified chronological splits contain temporal overlap, it is still not sufficient to improve model performance compared to random splits. This suggests that the temporal drift affects particular classes rather than the entire data set.

\section{Error Analysis}
\label{sec:erroranalysis}
Finally, we perform an error analysis to further investigate the type of errors made by BERT using both random and chronological splits. Table~\ref{tab:error} shows the number of correct and wrong predictions for each of the two data splitting strategies. We also use the Levenshtein distance\footnote{We set the threshold value to 20.} to calculate the quantity of posts in the test set that are similar to posts in the corresponding train set. 
    
    
\begin{itemize}
    \item We first observe that the temporal concept drift is evident in all rumor detection benchmarks. Most of the rumors on the same topic are posted in a very short time span.

    \item In addition, long-standing rumors are only a small part of the data (less than 5\%). Second, we note that using random splits leads to topical overlap between the training and test sets (see Table~\ref{tab:error_2}) resulting in  higher model performance.

    \item Finally, for both random and chronological splits, most of the posts in the test set with overlapping topics in the training set are predicted correctly. In contrast, wrong predictions are often posts with emerging or different topics compared to the posts in the train set.
\end{itemize}


\section{Conclusion}
We have shed light on the impact of temporal drift on computational rumor detection. Results from our controlled experiments show that the use of chronological splits causes substantially drops in predictive performance across widely-used rumor detection benchmarks. This suggests that random splits rather overestimate the model predictive performance. We argue that the temporal concept drift needs to be considered when developing real-world rumor detection approaches.
In the future, we plan to study the impact of temporal concept drift on other NLP tasks, such as detecting user reactions to untrustworthy posts on social media \citep{glenski-etal-2018-identifying,mu2020identifying,mu2023websci}.

\section*{Limitations}
We provide the first re-evaluation of four standard rumor detection benchmarks in two languages (English and Chinese) from two platforms (Twitter and Weibo). We acknowledge that further investigation is needed in rumor detection datasets in other languages. We provide an error analysis in Section~\ref{sec:erroranalysis}. 

\section*{Acknowledgments}
We would like to thank Ahmed Alajrami, Danae Sánchez Villegas, Mali Jin, Xutan Peng and all the anonymous reviewers for their valuable feedback.





\bibliography{anthology,GGWP}
\bibliographystyle{emnlp_natbib}





\end{document}